%% file: main.tex
\documentclass{article}

     \usepackage[preprint,nonatbib]{neurips_2020}
     
\usepackage{subcaption}
\usepackage{adjustbox}
\usepackage{amssymb}
\usepackage{amsmath}

\usepackage{multirow}
\usepackage{makecell}
\usepackage{framed}
\usepackage{stackengine}

\usepackage[utf8]{inputenc} 
\usepackage[T1]{fontenc}    
\usepackage{hyperref}       
\usepackage{url}            
\usepackage{booktabs}       
\usepackage{amsfonts}       
\usepackage{nicefrac}       
\usepackage{microtype}      
\newtheorem{definition}{Definition}

\usepackage{pgfplots}
\usepackage{tikz}

\usepackage[inline]{enumitem}
\newlist{inlinelist}{enumerate*}{1}
\setlist*[inlinelist,1]{label=\roman*),itemjoin={{; }},itemjoin*={{; and }}}

\input{math_commands.tex}

\title{Ranking Deep Learning Generalization using\\ Label Variation in Latent Geometry Graphs}

\author{%
  Carlos Lassance$^\dagger$
  \And
  Louis Béthune$^\circ$
  \AND
  Myriam Bontonou$^\dagger$
  \And 
  Mounia Hamidouche$^\dagger$
  \And 
  Vincent Gripon$^\dagger$
  \\ \vspace{0.5cm}
  $^\dagger$ IMT Atlantique, Brest, France \quad $^\circ$ Université Fédérale de Toulouse (ANITI), France

}
\begin{document}

\maketitle

\begin{abstract}
    Measuring the generalization performance of a Deep Neural Network (DNN) without relying on a validation set is a difficult task. In this work, we propose exploiting Latent Geometry Graphs (LGGs) to represent the latent spaces of trained DNN architectures. Such graphs are obtained by connecting samples that yield similar latent representations at a given layer of the considered DNN. We then obtain a generalization score by looking at how strongly connected are samples of distinct classes in LGGs. This score allowed us to rank 3rd on the NeurIPS 2020 Predicting Generalization in Deep Learning (PGDL) competition.
\end{abstract}

\section{Introduction}

Deep Neural Networks (DNNs) have achieved the state of the art in various machine learning tasks~\cite{he2016identity}. However, it is complicated to measure the performance of a trained DNN on data that has never been seen by the model during training. This is usually referred to as \emph{generalization}.

The most commonly used method to measure generalization involves splitting the training set to generate a validation one. The performance on the latter is used as a proxy to generalization~\cite{dlbook}. While this method works well in practice, it comes with the drawback of reducing the amount of training data and thus lowering the overall generalization performance of the network. Instead, being able to accurately measure the generalization of a network without reducing the amount of training data would be beneficial, in particular in data-thrifty scenarios.

Recently authors have tested various measures in order to solve this problem~\cite{Jiang2020Fantastic,bontonou2020predicting}, but it is still unclear what measures are robust to different types of networks and tasks~\cite{dziugaite2020search}. This paper introduces a new method that was proposed as a solution to the Predicting Generalization in Deep Learning (PGDL) competition held during the 2020 NeurIPS conference. The proposed method ranked 3rd out of 24 contributions.

The main idea of the proposed solution builds upon using Latent Geometry Graphs (LGGs)~\cite{lassance2020representing}. Such graphs are obtained by connecting samples of a batch depending on how similar their corresponding representations are. We then use label variation to determine how well this graph separates samples of distinct classes (via a metric called label variation). The obtained score indicates how well the latent space is aligned with the classification task to solve. Our results show that the proposed score correlates well with the generalization performance of considered architectures, independently of the training hyperparameters.

\section{Methodology}

\subsection{Latent Geometry Graphs}

We propose to study the latent geometries of DNNs by defining similarity graphs. In these graphs, vertices ($v \in \sV$) are data samples and the edge weight between two vertices depends on the similarity (e.g. cosine) between the corresponding intermediate representations at a given layer. We call such a graph a \textit{latent geometry graph (LGG)}. More details and motivation for the use of such representations are available in~\cite{lassance2020representing}. Let us detail how to build a LGG:
\begin{enumerate}
    \item Generate a symmetric matrix $\adjmatrix \in \R^{|\sV|\times |\sV|}$ using a similarity measure between intermediate representations, at a given depth $\ell$, of a batch of data samples $\mX$. In this work, we consider either cosine similarity or RBF similarity kernels.
    \item Threshold $\adjmatrix$ so that each vertex is only connected to its $k$-nearest neighbors.
    \item (Optional) Symmetrize the resulting thresholded matrix: two vertices $i$ and $j$ are connected with edge weights $w_{ij}=w_{ji}$ as long as one of the nodes was a $k$-nearest neighbor of the other one.
    \item (Optional) Normalize $\adjmatrix$ using its degree diagonal matrix $\mD$:  $\hat{\adjmatrix}= \mD^{-\frac{1}{2}} \adjmatrix \mD^{-\frac{1}{2}}$.
\end{enumerate}

\subsection{Label Variation}

Using the LGG associated with intermediate representations from a given layer $\ell$, we are able to measure the alignment between the representations of the data samples and the classification task under consideration. We measure this alignment using the label variation, a measure derived from the framework of Graph Signal Processing (GSP)~\cite{shuman2013emerging}. In a nutshell, this measure corresponds to the sum of all edge weights connecting samples of distinct classes. We borrow the motivation of using label variation (also called label smoothness) from~\cite{griort}. In the following paragraphs, we provide formal definitions for the label signal and the label variation:

\begin{definition}[Label signal]\label{label signal}
  The label signal is defined as the class indicator matrix (one-hot encoding)  of each sample $\mY\in\{0,1\}^{|\sV|\times|\sC|}$, where $\sC$ is the set of class labels.  
\end{definition}

\begin{definition}[Label variation]\label{label_variation}
Consider a LGG with adjacency matrix $\adjmatrix$ and the label signal $\mY$.  Label variation is defined as:
\begin{equation}
       \sigma_\ell = tr(\mY^{\top} \mL^\ell \mY)
       = {\sum_{i \in \sV} \sum_{j \in \sV}} {\adjmatrix^{\ell}_{i,j} \sum_{c \in \sC} \left(\mY_{i,c} - \mY_{j,c}\right)^2\;,}
\end{equation}
where $\mL^\ell=\mD^\ell - \adjmatrix^\ell$ is the combinatorial Laplacian of the LGG from layer $\ell$ and $\mD^\ell$ its degree matrix.

\end{definition}

Note that a small value of $\sigma$ indicates that the graph structure is well aligned with the classification task. However, there is a caveat that highly overfitted scenarios may also lead to small values of $\sigma$. 


\subsection{\emph{Mixup} augmented inputs}

In order to mitigate the risks of highly overfitted scenarios, we consider not only the examples of the training set, but also their augmentation using \emph{mixup}~\cite{zhang2017mixup}. The \emph{mixup} augmentation strategy simply consists in interpolating pairs of examples in the input space ($\mX_i,\mX_j$) and in the label space ($\mY_i, \mY_j$) using an interpolation factor $\lambda \sim \text{Beta}(\alpha, \alpha)$.

\subsection{Proposed approach}

For the PGDL submission, we use \emph{mixup} augmented samples to generate our LGGs and then, we use the label variation as our generalization measure. We generate $|\gG|$ graphs and consider our score to be the median score over these graphs. For each graph we sample $\max(1,\frac{500}{C})$ samples per class so that each graph has $|\sV| = \max(500,C)$ vertices.

We tried various combinations of LGGs, label variation and \emph{mixup} for the PGDL competition.  In this work we describe three of them: \begin{enumerate}
    \item \textbf{Variation Rate (VR)}: The average rate of change in label variation between the last three layers: $
        \frac {| \sigma_{3} - \sigma_{2} | + | \sigma_{2} - \sigma_{1} |}{2}\;,$
    where $\sigma_{i}$ refers to the $i$-th layer in the architectures starting from the end. This score comes from the experiments described in~\cite{griort}.
    \item \textbf{Worst Case Variation (WCV)}: The maximum value of label variation over the last 3 layers.
    \item \textbf{Variation of Penultimate layer with \emph{Mixup} (VPM)}: In this case, the score is simply $\sigma_{2}$. This is the score we have used on our final submission. 
\end{enumerate}

For all scores, different normalization techniques are applied in order to ensure that they are are comparable even when the graph size and the amount of connections can vary. We summarize the hyperparameters for each solution in Table~\ref{table:hyperparams}. We note that we tested many more hyperparameters, but for brevity we present only the ones with the best results.

\begin{table}[ht]
\caption{Table summarizing the hyperparameters for LGG construction for each score.}\label{table:hyperparams}
\adjustbox{max width=\columnwidth}{
\begin{tabular}{r|cccccc|c}
Measure                                                       & $|\gG|$ & Score & $k$  & Binarize after $k$-nn & Symmetrize & Normalize graph & $\alpha$ \\ \hline
VR                                               & 11  & Cosine & 20 & No                 & Yes        & No              & N/A   \\
WCV                                         & 1   & RBF    & 1  & No                 & Yes        & Yes             & N/A   \\
VPM                    & 80  & RBF    & 1  & Yes                & No         & Yes             & 2.0   \\
VPM (Final submission) & 1   & RBF    & 1  & Yes                & No         & Yes             & 2.0  
\end{tabular}
}
\end{table}

\section{Results on PGDL}

The PGDL competition had three different datasets: public, development and final. Each dataset had at least 2 distinct tasks, with the final dataset having 4 tasks. Competitors had access to the full public set, while the generalization data for the development and final sets were hidden. Evaluations for the development set were limited, but scoring feedback was given. For the final set scoring feedback was hidden until the end of the competition. 

Notably, in the final phase, tasks had more classes than in the public and development ones, leading to an increase in complexity for all solutions. In order to mitigate this problem, we had to reduce the number of graphs we consider, which led to less robust evaluations. On the other hand, this was probably the reason that we gained several positions on the final ranking compared to the development one. The code to reproduce our results on the public dataset is available at: \url{https://github.com/cadurosar/pgdl}. 

\subsection{Results on the public and development set}

Our results on the public and development sets are displayed in Table~\ref{tab:results_public}. At the start of the challenge we mostly concentrated in solutions based on the VR and WCV scores, without the use of \emph{mixup}. These solutions obtained great scores on the training set, with a pick at 32 on the public set. On the other hand, both of these scores were not able to generalize well to the development set. In order to improve generalization between sets we added \emph{mixup} augmentation to the training examples and reduced the complexity of our measure, leading to the VPM score.

\begin{table}[ht]
\centering
\caption{Results on the public and development sets. We were not able to test our final submission on the development set as the evaluation servers were closed.}\label{tab:results_public}
\adjustbox{max width=\columnwidth}{
\begin{tabular}{r|cc|cc|cc}
Measure     & Public   & Development & Task1 Public & Task 2 Public & Task 4 Dev & Task 5 Dev \\ \hline
VR          & 14.45    & 0.72        & 9.31         & 19.58         & 0.44       & 1.00       \\
WCV         & \textbf{32.6} & 0.37        & \textbf{27.74}        & \textbf{37.44}         & 0.21       & 0.55       \\
VPM         & 11.22    & \textbf{13.04}       & 5.61         & 16.82         & \textbf{15.42}      & \textbf{10.66}      \\ \hline
VPM (Final) & 6.26     & -           & 6.07         & 6.44          & -          & -         
\end{tabular}
}
\end{table}

The VPM scores allowed us to get results that are more balanced between the public and development sets, at the cost of a slightly lower performance. As expected, increasing $|\gG|$ led to better results on VPM, but due to the complexity of the final task we were only able to compare on the public set. Note that finding a balance between the three proposed scores (VPM/VR/WCV) and their hyperparameters should lead to better results overall, which is left as future work.

\subsection{Final results}

For the final evaluation, competitors could only send their solutions and be informed if they had finished in time or not. Due to these constraints we had to reduce the number of graphs to 1 in order to ensure that our solution would be able to run. On most tasks of the final set, our VPM (Final) ran very quickly. Indeed for the tasks 7, 8 and 9, we respectively used $7\%$, $1\%$, $1\%$ of the total allocated time, which means that we could easily increase the number of graphs we consider in order to get more accurate results. On the other hand, for task 6, we used $46\%$ of the total allocated time, which means that a trade-off between the number of graphs and the total amount of time is needed. 

Using only one graph for VPM resulted in a very large loss of performance in the public set as shown in the previous section. Otherwise, the results were pretty consistent over the public tasks. In Table~\ref{table:final_results} we present the per task results on the final set. The fact that there is a high variability on the results show that we should both use more graphs (in order to try to mitigate this variance) and better understand what is causing this variation. We also note that there is a large difference in score between the results obtained by the first place and the second/third (ours).

\begin{table}[ht]
\centering

\caption{Results on the final submission, we also present the first and second place mean results.}\label{table:final_results}
\begin{tabular}{c|c|ccccc}
Measure     & Mean on final set & Task 6 & Task 7 & Task 8 & Task 9 \\ \hline
First Place & \textbf{22.92}  & - & - & - & - \\
Second place & 10.16 & - & - & - & - \\ \hline
\textbf{Third Place - Ours (VPM Final)} & 9.99     & 13.90  & 7.56   & 16.23  & 2.28  \\ 
\end{tabular}
\end{table}

\section{Conclusion}

In summary, we have presented a solution that uses latent geometry graphs (LGGs) to capture the underlying geometry of the latent space and the label variation in order to create a proxy measure for generalization. This solution allowed us to get the 3rd place in the NeurIPS 2020 PGDL competition.

While we have presented our results here, it is hard to create a concise conclusion as a more in-depth ablation is needed to identify which parts were the most important for our solution. All in all, we were able to demonstrate the efficiency of our framework in measuring the generalization, but further study is still needed to understand which parts are the most important. Future work includes: \begin{enumerate}
\item Improving the sampling to generate our graphs (so that they concentrate on the most important pairs of samples of the training set), as proposed in~\cite{girault2020graph}. \item Improving the graph construction itself (how to correctly measure similarity in the latent space and to construct an efficient graph). We refer the reader to~\cite{lassance2020graph,shekkizhar2020graph} for a few examples. \end{enumerate}

\begin{ack}
Carlos Lassance was in part financed by the Brittany region of France. Louis is a PhD student under ANITI funding. Experiments on the public set used GPUs gifted by NVIDIA. We would like to thank Antonio Ortega for the invaluable discussions on LGGs and GSP, and also thank the organizers of the PGDL competition for their handling of the competition and the access to the computational cluster.
\end{ack}


\bibliography{main.bib}
\bibliographystyle{abbrv}

\end{document}

%% file: math_commands.tex
\usepackage{bm}

\def\eqref#1{equation~\ref{#1}}

\def\1{\bm{1}}

\def\mD{{\bm{D}}}

\def\mL{{\bm{L}}}

\def\mX{{\bm{X}}}
\def\mY{{\bm{Y}}}

\DeclareMathAlphabet{\mathsfit}{\encodingdefault}{\sfdefault}{m}{sl}
\SetMathAlphabet{\mathsfit}{bold}{\encodingdefault}{\sfdefault}{bx}{n}

\def\gG{{\mathcal{G}}}

\def\adjmatrix{{\mathbf{\mathcal{A}}}}

\def\sC{{\mathbb{C}}}

\def\sV{{\mathbb{V}}}

\newcommand{\R}{\mathbb{R}}

